%% file: paper.tex
\documentclass{ecai} 

\usepackage{latexsym}
\usepackage{amssymb}
\usepackage{amsmath}
\usepackage{amsthm}
\usepackage{booktabs}
\usepackage{enumitem}
\usepackage{graphicx}
\usepackage{color}

\usepackage{bm}
\usepackage{amsmath}
\usepackage{amssymb}
\usepackage{graphicx}
\usepackage{diagbox}
\usepackage{subfigure}
\usepackage{tikz}
\usetikzlibrary{shapes.geometric}
\usepackage{dblfloatfix}

\usepackage{xcolor}

\newcommand{\BibTeX}{B\kern-.05em{\sc i\kern-.025em b}\kern-.08em\TeX}

\begin{document}


\begin{frontmatter}


\paperid{123} 


\title{Rethinking Robustness: A New Approach to\\ Evaluating Feature Attribution Methods}


\author[A]{\fnms{Panagiota}~\snm{Kiourti}}
\author[B]{\fnms{Anu}~\snm{Singh}}
\author[B]{\fnms{Preeti}~\snm{Duraipandian}}
\author[A]{\fnms{Weichao}~\snm{Zhou}}
\author[A]{\fnms{Wenchao}~\snm{Li}}

\address[A]{Boston University, \{pkiourti, zwc662, wenchao\} @bu.edu}
\address[B]{Intuit Inc.}


\begin{abstract}
\input{sections/abstract.tex}
\end{abstract}

\end{frontmatter}


\begin{abstract}
    \input{./sections/abstract.tex}
\end{abstract}

\section{Introduction}
\label{introduction}
\input{./sections/introduction.tex}

\section{Background}
\label{background}
\input{./sections/background.tex}

\section{Related Works}
\label{related-work}
\input{./sections/related-work.tex}

\section{Approach}
\label{method}
\input{./sections/method.tex}

\section{Experiments}
\label{results}
\input{./sections/results.tex}
\section{Conclusion}
\label{conclusion}
\input{./sections/conclusion.tex}

\bibliography{paper}

\end{document}

%% file: sections/abstract.tex
This paper studies the robustness of feature attribution methods for deep neural networks. It challenges the current notion of attributional robustness that largely ignores the difference in the model's outputs and introduces a new way of evaluating the robustness of attribution methods. Specifically, we propose a new definition of \textit{similar} inputs, a new robustness metric, and a novel method based on generative adversarial networks to generate these inputs. In addition, we present a comprehensive evaluation with existing metrics and state-of-the-art attribution methods. Our findings highlight the need for a more objective metric that reveals the weaknesses of an attribution method rather than that of the neural network, thus providing a more accurate evaluation of the robustness of attribution methods. 

%% file: sections/introduction.tex
Explainability is crucial to building trust in AI systems and evaluating their safety. Feature attribution methods are one approach for explaining a deep learning model's behavior by assessing how much each input component contributes to the model's prediction~\cite{simonyan2013deep,zeiler2014visualizing,springenberg2014striving,bach2015pixel,mahendran2016salient,shrikumar2016not,sundararajan2017axiomatic,smilkov2017smoothgrad,shrikumar2017learning}. 
However, it is difficult to assess the quality of an attribution method as ground-truth attributions are often not available. 

The need to evaluate explainability methods has inspired research to define a set of desired properties for attributions. Fidelity or faithfulness~\cite{adebayo2018sanity,bhatt2020evaluating,velmurugan2021evaluating}, and robustness or sensitivity~\cite{ancona2017towards,ghorbani2019interpretation,bhatt2020evaluating,hsieh2020evaluations,kamath2022robust,fel2022good,agarwal2022rethinking} are the most studied properties. In particular, existing methods consider an attribution robust if a small perturbation in the input does not substantially change the explanation~\cite{alvarez2018robustness,ghorbani2019interpretation,dombrowski2019explanations,sarkar2021enhanced,agarwal2022rethinking}. 
This concept is usually referred to as \textit{attributional robustness}. 
However, small perturbations in the input are also used to evaluate adversarial robustness of neural networks~\cite{szegedy2013intriguing,goodfellow2014explaining,carlini2017towards,madry2017towards}. 
As a result,  approaches aimed at evaluating attributional robustness often borrow techniques from adversarial robustness to generate these perturbations.
Other approaches use the training dataset to find a neighborhood around an input $\bm{x}$ and evaluate the robustness of input's $\bm{x}$ attribution by comparing it against attributions from the inputs in the neighborhood of $\bm{x}$~\cite{bhatt2020evaluating}.

The existing methods implicitly make one or more of the following assumptions when evaluating attributional robustness: (a) inputs are considered similar by the model if they are numerically or visually close, (b) the given training dataset can be used to define the neighborhood of new inputs, that is, the model's learned input distribution is aligned with the training dataset distribution perfectly, and the dataset has no impurities or bias~\cite{casper2023red}, and (c) the model is robust against small perturbations in the sense that it does not change its output which is also used in the computation of attributions. We argue that similar inputs generated under assumptions (a), (b), or (c) are not necessarily considered similar according to what the model has learned. 

In this paper, we call attention to the importance of considering the output differences of a neural network when identifying input similarity for attributional robustness and propose a new definition for such analysis. As shown in~\cite{kamath2022robust}, existing metrics that evaluate the current notion of attributional robustness are sensitive to random changes in the input that do not accurately reflect the robustness of attribution methods. Metrics that rely on visually similar inputs alone to evaluate attributional robustness implicitly introduce into their evaluation the bias of defining input similarity as visual or numerical similarity and how well the model is trained to recognize such inputs as similar. 
Thus, we argue that the robustness of an attribution method should be \textit{independent} from how well a model is trained to a given training distribution and, hence, independent from the adversarial robustness of the model.
    


In other words, we challenge what it means for two inputs to be \textit{similar} in the context of whether they produce similar explanations. We aim to decouple the notion of adversarial robustness from the notion of attributional robustness. 
Our contributions are summarized below.

\begin{enumerate}
    \item We propose a new definition of similar inputs for evaluating attributional robustness.
    \item We introduce Output Similarity-based Robustness (OSR), a robustness metric that challenges the previous definition of attributional robustness which is intuitively based only on small perturbations in the input.
    \item We develop a novel method that uses Generative Adversarial Networks (GANs) to synthesize similar inputs for evaluating OSR. 
    \item We comprehensively evaluate our metric against existing metrics and show that OSR is more suitable for capturing attributional robustness.
\end{enumerate}

%% file: sections/background.tex
\subsection{Attributions} 
In machine learning models, feature attribution methods are used to compute the contribution of each input component (feature) $\bm{x}_i$ of a given input $\bm{x} \in \mathcal{R}^d$ to the prediction $y = f(\bm{x})$ of a given model $f$. Throughout the paper, we use $g(\bm{x})$ to refer to the attributions of an input $\bm{x}$ and $g$ to refer to the attribution's method function. In particular, $g(\bm{x})$ is a vector $\bm{a} \in \mathcal{R}^d$, where each $\bm{a}_i$ is a number representing the contribution of $\bm{x}_i$ towards the prediction. 

\subsection{Attributional Robustness} 
\textit{Attributional robustness} was first introduced through adversarial attacks towards attributions~\cite{ghorbani2019interpretation,dombrowski2019explanations,zhang2020interpretable}, often called attributional attacks~\cite{kamath2022robust}. Given a model $f$ and an input $\bm{x}$ with predicted class $y = f(\bm{x})$, attributional attacks introduce an input $\widetilde{\bm{x}}$ that is predicted as $y$ and is similar to $\bm{x}$ based on an $\ell_p$ norm but exhibits attributions $g(\widetilde{\bm{x}})$ that have large differences from the original input's attributions $g(\bm{x})$, despite the similarity in the input space. According to these works, comparing the attributions $g(\bm{x})$ of the original input to the attributions $g(\widetilde{\bm{x}})$ can reveal the attributional robustness around $\bm{x}$ for the attribution method $g$ and the model $f$. However, as mentioned in Section~\ref{introduction}, the current works~\cite{dombrowski2019explanations,ghorbani2019interpretation,casper2023red} introduce similar inputs using concepts from the field of adversarial robustness and, therefore, introduce the bias of the adversarial robustness of the model. 

At the same time,~\cite{bhatt2020evaluating} and~\cite{hsieh2020evaluations} introduced desired properties and evaluation metrics for attribution methods. Both the attributional attack papers and the works that introduce evaluation metrics for attributions describe two steps. The first step finds similar inputs $\widetilde{\bm{x}}$ around an input $\bm{x}$, while the second step compare the attributions $g(\bm{x})$ and $g(\widetilde{\bm{x}})$. Current methods use the $\ell_1$~\cite{zhang2020interpretable}, or $\ell_2$ norm ~\cite{bhatt2020evaluating,hsieh2020evaluations} to compare attributions, but they also borrow ranking metrics from other fields. In particular, top-$k$ intersection is used to find the percentage of the top-$K$ attributed input features that are the same between the two attributions as well as Kendall's $\tau$ or Spearman's $\rho$ correlation between the features of the two attribution maps~\cite{ghorbani2019interpretation,sarkar2021enhanced,agarwal2022rethinking,kamath2022robust}. However, as mentioned in~\cite{agarwal2022rethinking}, those metrics might show low robustness because they unfairly penalize attribution changes. In what follows, we introduce the robustness and evaluation metrics that are well-known in the field of XAI.

\textit{Sensitivity} is defined by~\cite{bhatt2020evaluating} as the $\ell_2$ distance of the attribution of $\bm{x}$ from the attributions of the training points that are $\ell_\infty$-close to $\bm{x}$ and belong to the predicted class: $\mu_M = \max_{\bm{x}^\prime\in \mathcal{N}_r} \|g(\bm{x}) - g(\bm{x}^\prime)\|_{2}$, where $\mathcal{N}_r = \{\bm{x}^\prime \in \mathcal{R}^d \mid \|\bm{x} - \bm{x}^\prime\|_{\infty} \leq \rho, f(\bm{x}) = f(\bm{x}^\prime)\}$. In other words, sensitivity expects that inputs close to each other in terms of the $\ell_\infty$ norm have close attributions in terms of the $\ell_2$ norm. The lower the sensitivity, the more robust the attributions around $\bm{x}$. Based on this definition, sensitivity is another word for attributional robustness. 

\textit{Fidelity} is a popular evaluation metric that aims to assess whether the attribution method correctly identifies the features that the neural network relies on to make a decision~\cite{adebayo2018sanity,bhatt2020evaluating}.~\cite{bhatt2020evaluating} defined fidelity as the correlation of the sum of attributions of different sets of input features to the output difference when those sets of features are removed: 

$$\mu_F = \texttt{corr}_{S\in {d \choose \mid S\mid}} \left( \sum_{i\in S} g(\bm{x})_i, f(\bm{x}) -f(\bm{x}_{[\bm{x}_s = \bm{x}_b]})\right)$$

\noindent where $\bm{x}_b$ is a baseline image that is expected to produce $f(\bm{x}_b) \approx 0$, $S$ is a subset of input features in the combination set $d \choose \mid S\mid$, and $d$ is the number of input features. The size of $S$ is set to be a fixed number in practice for simplicity. 
Intuitively, this metric computes the correlation between the importance of different sets of features and the difference in the output logits when we remove those features by setting them to baseline values $\bm{x}_b$. Hence, the higher the fidelity is, the more reliable the attribution method. 

Finally,~\cite{hsieh2020evaluations} introduced \textit{robustness}-$S_r$ as an evaluation metric for attributions. Robustness-$S_r$ is defined as the minimum $\ell_2$-norm perturbation $\delta$ on a set of features $S$ that can cause misclassification for a given input $\bm{x}$, i.e., $S_r = \{\min_{\bm{\delta}} \|\bm{\delta}\|_p \; s.t. \; f(\bm{x} + \bm{\delta}) \neq y\}$, where $\bm{\delta}$ is a vector that introduces perturbations only to the input features that belong to a pre-selected set $S$. The smaller the robustness $S_r$ is, the more robust the attribution method is. 


%% file: sections/related-work.tex
The notion of attributional robustness has been introduced over the past few years as a metric to evaluate how sensitive an attribution method is around an input $\bm{x}$ when certain input changes are introduced and, therefore, how good an attribution method is for a specific dataset and model~\cite{yeh2019fidelity,bhatt2020evaluating,agarwal2022rethinking}. Most of these works have focused on the network's gradients with respect to the input, Lipschitz continuity of the model, or adversarial robustness~\cite{alvarez2018robustness,chen2019robust,agarwal2022rethinking,hsieh2020evaluations,singh2020attributional} to define how sensitive an attribution method is to input perturbations. Other works~\cite{ghorbani2019interpretation,chen2019robust,huang2023robustness} use ranking metrics, such as top-$k$ intersection, Spearman's $\rho$ or Kendall's $\tau$ correlation coefficient, to evaluate the sensitivity of attributions by comparing the ranking of the input components that results from the attribution method. 


~\cite{hsieh2020evaluations} introduced both a robustness metric, called \textit{Robustness}-$S_r$, and the attribution method GreedyAS that optimizes this metric. Robustness-$S_r$ is an evaluation metric defined as the minimum perturbation on an input $\bm{x}$ in terms of an $\ell_p$-norm that causes the model to misclassify the perturbed input. Since GreedyAS optimizes robustness $S_r$, it is expected that GreedyAS will exhibit the highest robustness when Robustness-$S_r$ is used to evaluate attribution methods. Therefore, using $S_r$ introduces bias in the evaluation of the methods. 

Other works around attributional robustness focus on adding regularization terms in the training loss function to ensure that the resulting model produces robust attributions for a given input $\bm{x}$~\cite{kamath2022robust,sarkar2021enhanced}.

Our paper falls into the category of attributional robustness for evaluating current attribution methods on a specific task. We are the first to define the notion of attributional robustness to use similar inputs according to what the model identifies as similar instead of relying on adversarial attacks, which introduces a dependency on the adversarial robustness of the model.

%% file: sections/method.tex



\subsection{Redefining Similar Inputs}
We first introduce our definition of similar inputs.

\paragraph{Definition 1.}  Given an image $\bm{x}\in \mathcal{R}^d$, a classifier $f$, predicted class $y$, and the scalar value $F_y(\bm{x})$ which is the value of the logit for the predicted class $y$, we define an input $\bm{\widetilde{x}}$ to be similar to $\bm{x}$ if $\widetilde{\bm{x}}$ belongs to a set $\mathcal{S}$, where

\begin{equation}
\label{similar_inputs_definition}
\mathcal{S} = \left\{\widetilde{\bm{x}} \in \mathcal{R}^d \; \Big| \; |F_y(\bm{x}) - F_y(\widetilde{\bm{x}})| \leq \delta, \;\| \bm{\widetilde{x}} - \bm{x} \|_2 \leq \rho\right\}.
\end{equation}
The constants $\delta$ and $\rho$ represent bounds on the input and output differences, respectively, to measure similarity both in the input and output space for a new input fed through the model. From hereafter we refer to $F_y(\bm{x})$ as the prediction logit. Hence, $\delta$ is the distance between the prediction logit of input $\bm{x}$ and the prediction logit of input $\widetilde{\bm{x}}$, that is, $\left| F_y(\bm{x}) - F_y(\widetilde{\bm{x}})\right|$. 
$\rho$ represents the maximum $\ell_2$ distance of the new input $\widetilde{\bm{x}}$ from $\bm{x}$ and aims to limit how far from the original input, the new inputs can be, in order to enforce similarity in the input space.

\subsection{Problem Definition}
In this paper, we solve the problem of evaluating the different attribution methods objectively on a given distribution $\mathcal{X}$ regardless of how well the model is trained on $\mathcal{X}$ or its adversarial robustness. 

Given a model $f$, an input $\bm{x}\sim\mathcal{X}$, and a prediction $y = f(\bm{x})$, we define a neighborhood of $\bm{x}$ that is not biased by the model's adversarial robustness in order to evaluate the robustness of different attribution methods $g$. In what follows, we consider an attribution method to be robust around an input $\bm{x}$ if the attributions $g(\bm{x})$ are numerically close to the attributions of inputs that exist around a neighborhood of $\bm{x}$ and are classified as $y$ with prediction logit close to the prediction logit of input $\bm{x}$.

\subsection{Redefining Attributional Robustness}
\paragraph{Definition 2.} Given a classifier $f$, an input $\bm{x} \in \mathcal{R}^d$, and logits $F(\bm{x})$ we introduce Output Similarity-based Robustness (OSR), a robustness metric that evaluates if an explanation function $g$ is $\epsilon,\delta$-\textit{robust} at $\bm{x}$ by computing how close the attributions $g(\bm{x})$ are to the attributions on inputs similar to $\bm{x}$ according to Definition~\ref{similar_inputs_definition}. In particular, $g$ is $\epsilon,\delta$-\textit{robust} at $\bm{x}$ if:
\begin{equation}
OSR(\bm{x}) = \mathbb{E}_{\widetilde{\bm{x}}\sim D}\left[\lVert g(\bm{x}) - g(\widetilde{\bm{x}})\rVert_2\right]\leq \epsilon,
\end{equation}

\noindent where $\mathcal{S}$ is defined in Eq.~(\ref{similar_inputs_definition}), $\delta$ and $\epsilon$ are small numbers, and $D$ is the unknown true distribution that represents the set $S$.

Intuitively, our robustness metric is small when $g$ produces $\epsilon$-close explanations regarding the $\ell_2$ norm for the distribution of similar inputs that have $\delta$-close prediction logits to the original input's $\bm{x}$ prediction logit, i.e., $\left|F_y(\bm{x}) - F_y(\widetilde{\bm{x}}))\right| \leq \delta$. That is, $g$ is robust when it produces \textit{similar explanations} for inputs that the neural network considers \textit{similar} regarding the assigned logits of the predicted class $y$. In this case, the smaller the $OSR(\bm{x})$, the more robust $g$ is around $\bm{x}$. 

\subsection{Motivating Examples for the New Metric}
\label{sec:motivating}
This section presents examples that motivate the need for a better definition of similar inputs than solely relying on inputs with minor perturbations. 

We introduce a Generative Adversarial Network (GAN) to approximate the unknown distribution $D$ of similar images (the set $S$ in Definition~\ref{similar_inputs_definition}). The following figures will show how existing, more straightforward techniques fall short in approximating this unknown distribution because they are not designed for objective evaluation of attributional robustness. In particular, the existing techniques either add noise to the given image $\bm{x}$~(\cite{ghorbani2019interpretation}), sampled from a distribution with parameters chosen arbitrarily, or leverage adversarial methods to search for inputs in a chosen $\ell_p$ ball around $\bm{x}$. Both approaches require choosing values for parameters that introduce selective bias. 

\begin{figure}
    \centering
    \subfigure[][]{
        \label{fig:eval-a}
        \includegraphics[height=1.05in]{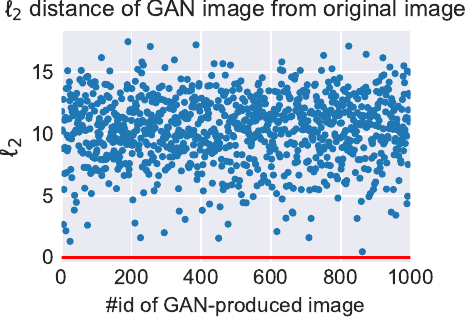}
    }
    \subfigure[][]{
        \label{fig:eval-b}
        \includegraphics[height=1.05in]{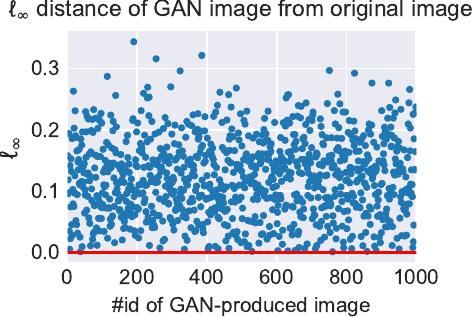}
    }
    \subfigure[][]{
        \label{fig:eval-c}
        \includegraphics[height=1.05in]{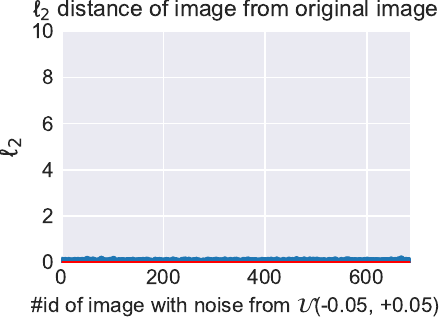}
    }
    \subfigure[][]{
        \label{fig:eval-d}
        \includegraphics[height=1.05in]{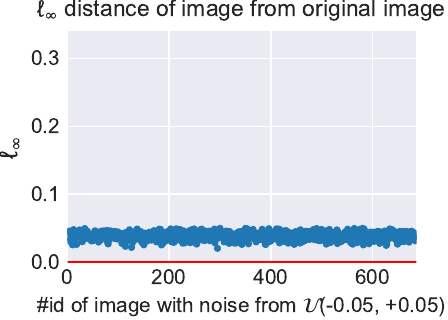}
    }
    \subfigure[][]{
        \label{fig:eval-e}
        \includegraphics[height=1.05in]{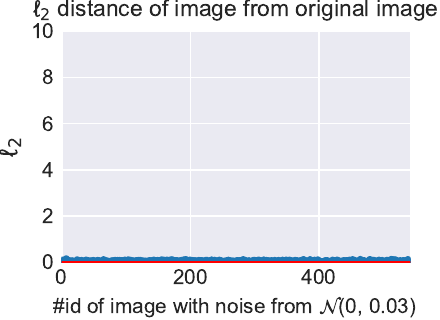}
    }
    \subfigure[][]{
        \label{fig:eval-f}
        \includegraphics[height=1.05in]{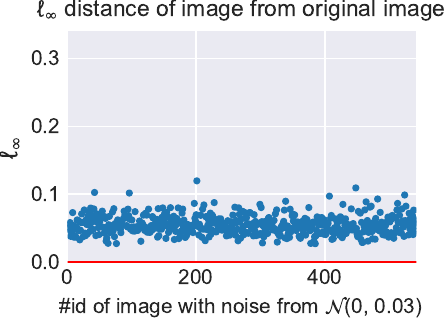}
    }
    \caption[Distances of generated images from the original image $\bm{x}$ for: (a), (b) GAN-generated images; (c),(d) images with uniformly sampled noise; (e),(f) images with noise from $\mathcal{N}(0,0.05)$.]{Distances of generated images from the original image $\bm{x}$ for: (a), (b) GAN-generated images; (c),(d) images with uniformly sampled noise; (e),(f) images with noise from $\mathcal{N}(0,0.03)$.}
    \label{fig:osr}
\end{figure}

In Fig.~\ref{fig:eval-c} and ~\ref{fig:eval-d}, we show how adding noise from a uniform distribution $\mathcal{U}(-0.05, 0.05)$ 
generates images with $\ell_p$ distances from the original image concentrated around a specific value. However, our approach reveals that this set of images, with a small $\ell_2$ distance to the original, does not encompass the entirety of similar images. Additionally, this method produces images that are $\sim50\%$ misclassified. By slightly increasing the noise to $0.06$, the misclassification rate rises to $85\%$, while still maintaining similarity to the original images and concentrated $\ell_p$ distances.  Similar outcomes are observed when employing a normal distribution $\mathcal{N}(0, 0.03)$ for noise generation (Fig.~\ref{fig:eval-e},~\ref{fig:eval-f}). Increasing the standard deviation to $0.04$ results in merely about $25$ correctly classified images out of 1000, despite their small $\ell_p$ distances from the original image.

\begin{figure}
    \centering
    \includegraphics[height=1.2in]{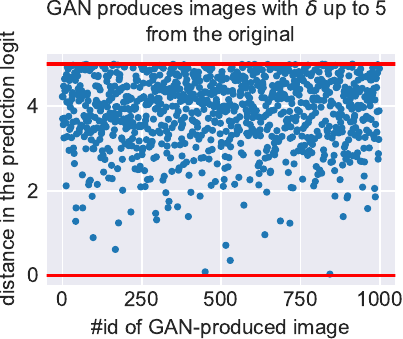}
    \caption{Distances between the prediction logit of the generated image and that of the original image.}
    \label{fig:deltas}
\end{figure}

In contrast, our GAN eliminates the need for bounds in the input space by incorporating closeness in the network's output space. Our GAN generates images with prediction logit that deviates by up to $\delta$ from $\bm{x}$'s logit (Fig.~\ref{fig:deltas}) while still maintaining visual and numerical similarity to the image $\bm{x}$. The $\ell_p$ distances in the input space are not concentrated around a value, as shown in Fig.~\ref{fig:eval-a} and~\ref{fig:eval-b}. The distances in the output space are also not exclusively concentrated near the boundary $\delta$ (Fig.~\ref{fig:deltas}). This observation persists even when generating fewer images (such as $50$). 


By increasing $\delta$, we can discover a broader set of images similar to $\bm{x}$. It is important to note that the generated images' prediction logit will not exceed a certain value, as this would lead to misclassification. This constraint is enforced by training the GAN to produce images with the same prediction as $\bm{x}$, limiting the extent of logit changes despite the use of a very large $\delta$. Hence, opting for a large $\delta$ ensures similar images within the maximum output distance that does not cause misclassification. Therefore, in our experiments, we choose a high $\delta$ value to ensure that we effectively approximate the set of similar images.

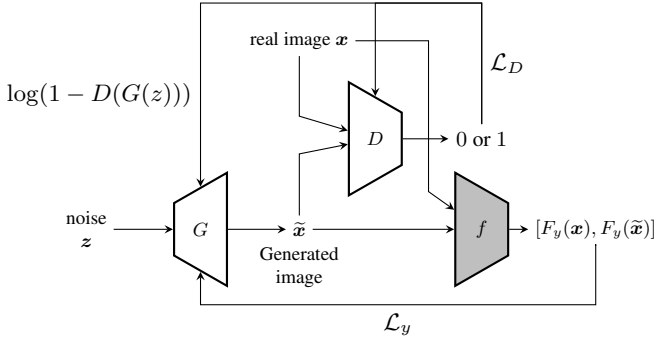
\begin{figure}[ht]
\begin{tikzpicture}
    \node[rectangle, scale=0.8] (real_image) at (2.8,0.5) {real image $\bm{x}$};
    \node[trapezium, trapezium angle=60, rotate=-90, minimum width=15mm, draw, thick] (discriminator) at (3.8,-0.8) {};
    \node[rectangle, scale=0.8] (label_d) at (3.8, -0.8) {$D$}; 

    \node[rectangle,scale=0.9] (label_dout) at (5.2, -0.8) {$0$ or $1$};
    
    \node[rectangle, scale=0.8,align=center] (noise) at (0.0, -2.0) {noise\\$\bm{z}$}; 
    
    \node[trapezium, trapezium angle=60, rotate=90, minimum width=15mm, draw, thick] (generator) at (1.5,-2.0) {};
    \node[rectangle, scale=0.8] (label_g) at (1.5, -2.0) {$G$}; 
    
    \node[rectangle, scale=0.8] (generated_image) at (2.8, -2.0) {$\widetilde{\bm{x}}$};
    \node[rectangle, scale=0.8,align=center] (label_gan_image) at (2.8, -2.5) {Generated\\ image}; 

    \node[trapezium, trapezium angle=60, fill=lightgray, rotate=-90, minimum width=15mm, draw, thick] (model) at (5.2,-2.0) {};
    \node[rectangle, scale=0.8] (label_f) at (5.2, -2.0) {$f$}; 
    \node[rectangle, scale=0.8] (label_fout) at (6.7, -2.0) {$[F_y(\bm{x}), F_y(\widetilde{\bm{x}})]$}; 
    
    \draw[->,>=stealth] (real_image) -- (2.8,-0.5) -- (discriminator);
    \draw[->,>=stealth] (noise) -- (generator);
    \draw[->,>=stealth] (generator) -- (generated_image);
    \draw[->,>=stealth] (generated_image) -- (2.8, -1.0) -- (discriminator);
    \draw[->,>=stealth] (generated_image) -- (model);
    \draw[->,>=stealth] (real_image) -- (4.5, 0.5) -- (4.5,-1.5) -- (model);
    \draw[->,>=stealth] (discriminator) -- (label_dout);
    \draw[->,>=stealth] (model) -- (label_fout);
    \draw[->,>=stealth] (label_fout) -- (6.7,-2.2) -- (6.7,-3.0) -- (1.5,-3.0)node[midway,below,align=center] {$\mathcal{L}_y$} -- (generator);
    \draw[->,>=stealth] (label_dout) -- (5.2,1.0) -- (1.5,1.0) -- (generator)node[midway,align=center,left] {$\log(1 - D(G(z)))$};
    \draw[->,>=stealth] (label_dout) -- (5.2,1.0)node[midway,right,align=center] {$\mathcal{L}_D$} -- (3.8,1.0) -- (discriminator);
    
\end{tikzpicture}
\caption{The proposed GAN-based method for synthesizing similar inputs. The neural network $f$ is already trained and has fixed parameters during the training of the GAN.}
\label{fig:gan}
\end{figure}

\subsection{Inputs with $\delta$-close Logits}
We propose a Generative Adversarial Network (GAN) to approximate the distribution of similar inputs that are close to input $\bm{x}$ with prediction logit $F_y(\widetilde{\bm{x}})$ at most $\delta$ away from the prediction logit $F_y(\bm{x})$. 

The original GAN architecture comprises a generator network $G$ and a discriminator network $D$. The generator is tasked with producing fake images from a noise vector $z$, aiming to resemble real images sampled from a true distribution $\mathcal{X}$. Meanwhile, the discriminator learns to differentiate between real and fake images by assigning labels of 1 to real images and 0 to fake ones. The original loss function of a GAN is formulated as a minimax optimization problem~\cite{goodfellow2014generative}:
\begin{align}
    \min_G \max_D V(D,G) =&\; \mathbb{E}_{\bm{x}\sim\mathcal{X}} \left [\log D(\bm{x}) \right ] \nonumber \\ 
    +&\; \mathbb{E}_{\bm{z}\sim P_{\bm{z}}(\bm{z})} \left [\log (1 - D(G(\bm{z}))) \right]
\end{align}

\noindent where the generator $G$ endeavors to produce images $G(\bm{z})$ that the discriminator $D$ perceives as real, denoted by $D(G(z))=1$. Conversely, the discriminator $D$ strives to learn to differentiate between the generated images $G(\bm{z})$ and the real images $\bm{x}$, specifically aiming for $D(G(\bm{z}))=0$ and $D(\bm{x})=1$.

In our setting, our objective is to ensure that the generated images closely resemble the original image while also having prediction logits $F_y$ within a distance of $\delta$ from the prediction logit of the original input $\bm{x}$. To approximate this distribution of inputs using a GAN, we incorporate an additional loss term. This term penalizes the distance between the prediction logits of the generated inputs and the prediction logits of the original input $\bm{x}$ when both are passed through the model $f$. Our additional loss term for the generator is:
\begin{equation}
    \mathcal{L}_y = 
    \begin{cases} 
    \lvert F_y(\bm{x}) - F_y(G(\bm{z})) - \delta \rvert, & F_y(\bm{x}) - F_y(G(\bm{z})) < \delta \\
    0, & F_y(\bm{x}) - F_y(G(\bm{z})) \geq \delta
    \end{cases}
\end{equation}

\noindent We denote the loss concerning the output logits as $\mathcal{L}_y$. To ensure that the images $\widetilde{\bm{x}}$
 are classified into the same class $y$, we incorporate a cross-entropy loss for the logits $F(\bm{x})$ and $F(\widetilde{\bm{x}})$. Additionally, we utilize multiple copies of $\bm{x}$ as the dataset for the real images, as we generate images corresponding to each original input $\bm{x}$ and train a distinct GAN for each input. Therefore, to obtain similar images as defined in Eq.~(\ref{similar_inputs_definition}), we optimize the following loss function for the generator:

\begin{equation}
    \mathbb{E}_{\bm{z}\sim P_{\bm{z}}(\bm{z})} \left [\log (1 - D(G(\bm{z}))) \right] + \mathcal{L}_y - F(G(\bm{z}))\log(F(G(\bm{z}))).
\end{equation}

\noindent The first term is the original loss of the generator, the second term is the loss regarding the output logits, and the last term is the cross-entropy loss to ensure that the generated images fall into the category that the original image is classified to. Finally, we keep the loss of the discriminator as it is. Intuitively, given an input $\bm{x}$, we move away from $\bm{x}$ so that the logits $F_y$ of the two inputs are at most $\delta$ apart and the inputs are classified to the same class. Our proposed GAN and the training loss terms are shown in Fig.~\ref{fig:gan}. $L_D$ denotes the loss of the discriminator as originally introduced by~\cite{goodfellow2014explaining} ($\log(1 - D(G(z)))$) while the generator's loss is enhanced with the term $L_y$ to enforce output similarity and the cross-entropy loss to enforce same prediction.


\begin{figure*}
    \centering
    \includegraphics[width=0.99\linewidth]{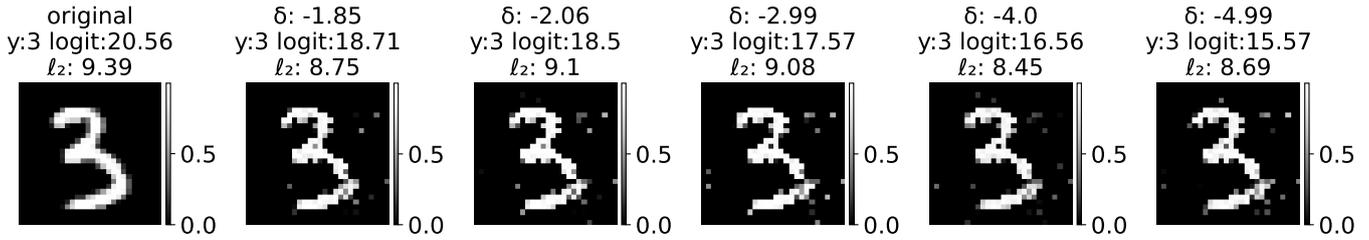}
    \caption{GAN-generated MNIST images with prediction logit \textit{at most} $\delta = 5.0$ far away from the prediction logit of the original image (on the left).}
    \label{fig:mnist_gan}
\end{figure*}

\begin{figure*}[ht]
    \centering
    \includegraphics[width=0.99\linewidth]{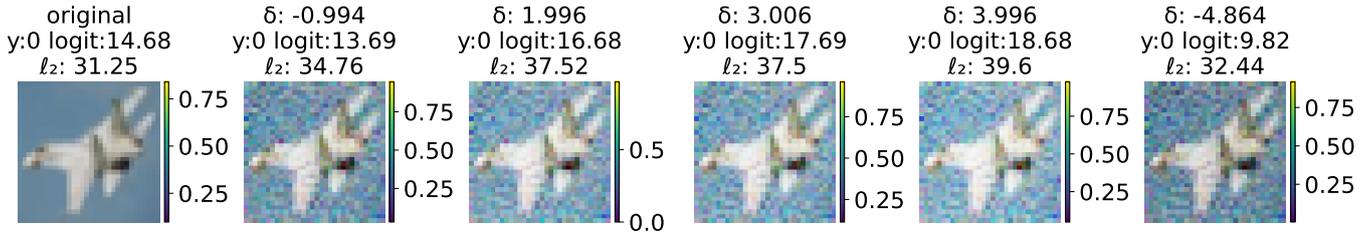}
    \caption{GAN-generated CIFAR10 images for the class 0 (airplane) with maximum logit \textit{at most} $\delta=5.0$ far away from the maximum logit of the original image (left).}
    \label{fig:cifar10_gan}
\end{figure*}

\begin{figure*}
    \centering
    \includegraphics[width=0.99\linewidth]{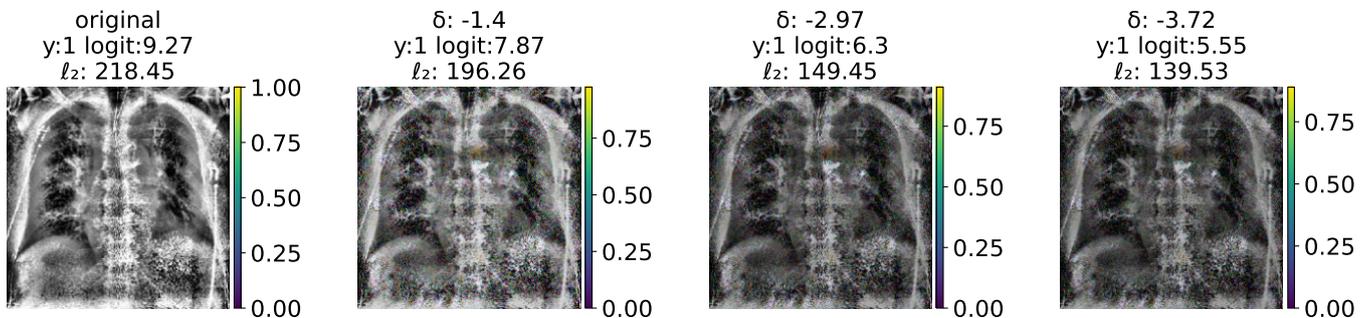}
    \caption{GAN-generated Chest-Xray images for the class 1 (COVID19) with prediction logit \textit{at most} $\delta=5.0$ away from the prediction logit of the original image (left).}
    \label{fig:covid_gan}
\end{figure*}

%% file: sections/results.tex
\begin{figure*}
    \centering
    \subfigure[][]{
        \label{fig:roc-a}
        \includegraphics[height=1.9in]{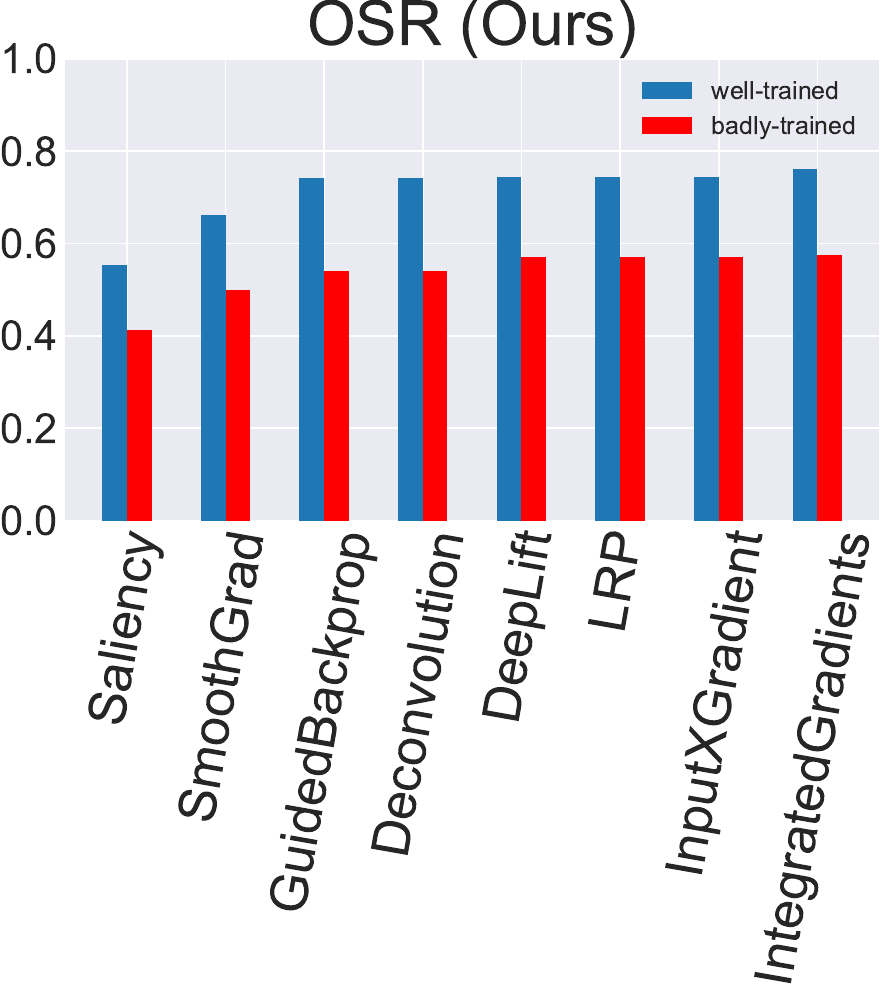}
    }
    \subfigure[][]{
        \label{fig:roc-c}
        \includegraphics[height=1.9in]{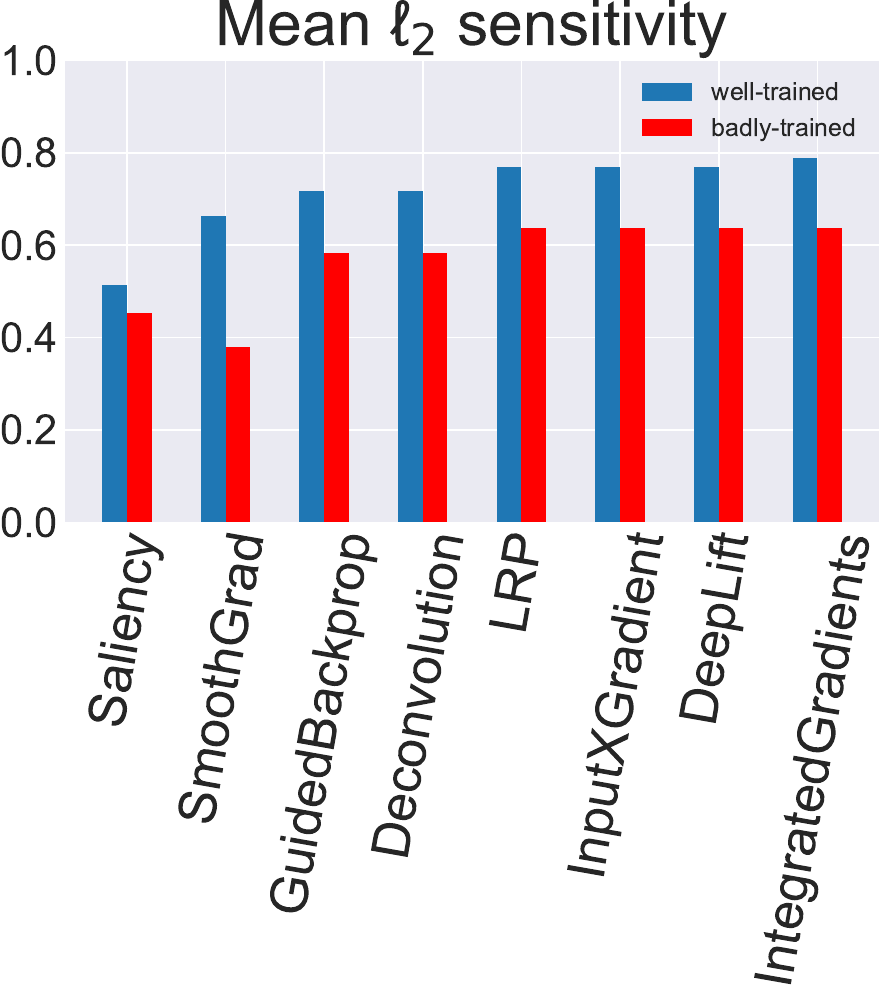}
    }
    \subfigure[][]{
        \label{fig:roc-d}
        \includegraphics[height=1.9in]{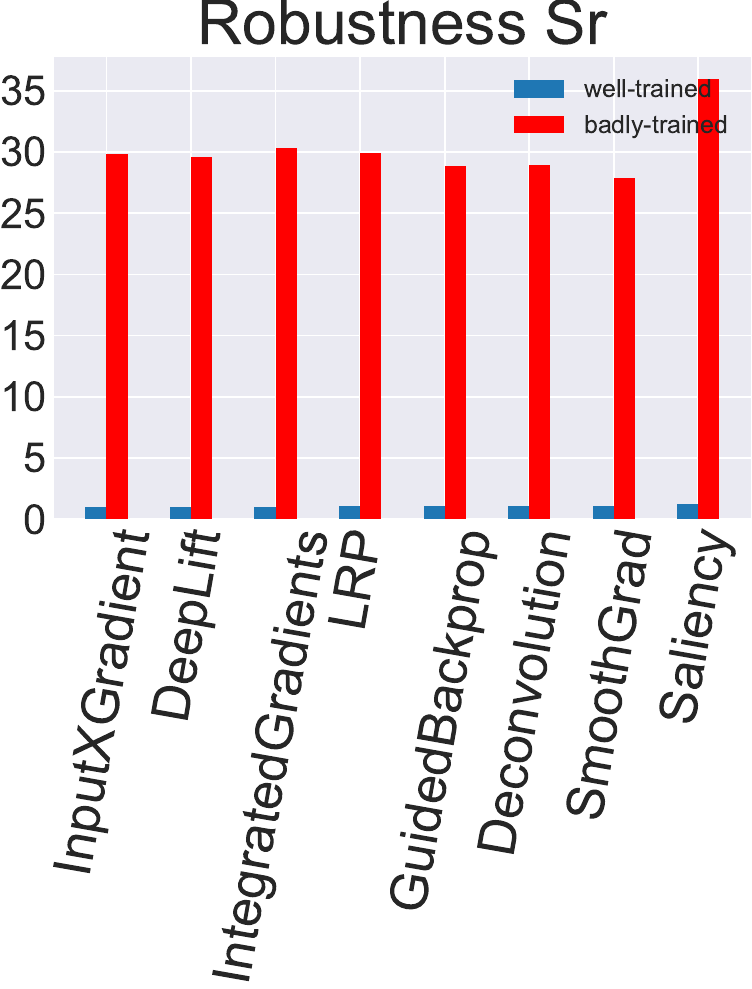}
    }
    \subfigure[][]{
        \label{fig:roc-e}
        \includegraphics[height=1.9in]{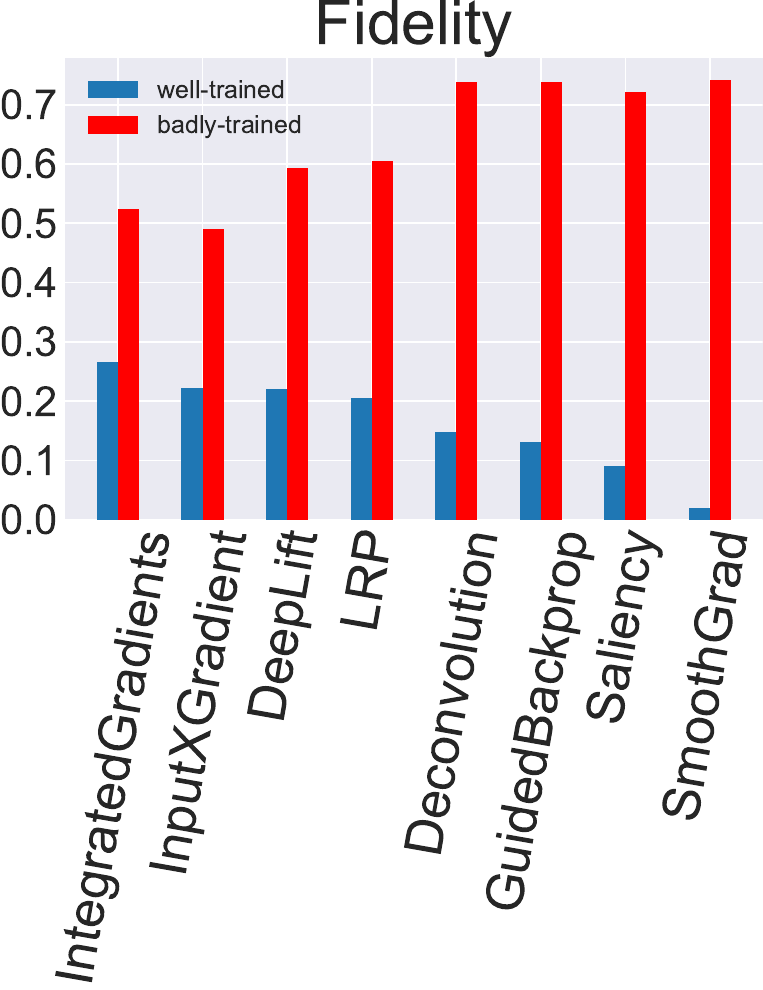}
    }
    \caption[Attributional Robustness and Fidelity.]{Comparison of attribution robustness metrics and fidelity on well-trained (\textbf{blue}) and badly-trained inputs (\textbf{red}) for MNIST:
    \subref{fig:roc-a} OSR (Ours): Lower is better;
    \subref{fig:roc-c} Mean $\ell_\infty$ sensitivity $\mu_A$: Lower is better;
    \subref{fig:roc-d} Robustness $S_r$: Lower is better;
    \subref{fig:roc-e} Fidelity $\mu_F$: Higher is better.}
    \label{fig:robustness_metrics_mnist}
\end{figure*}

The proposed robustness metric was evaluated on three datasets from the image domain: MNIST~\cite{deng2012mnist}, CIFAR10~\cite{krizhevsky2009learning}, and Covid19-Pneumonia-Normal Chest X-Ray Images Dataset~\cite{shastri2022cheximagenet,kumar2022litecovidnet}. We trained two models for each dataset -- one with a high accuracy and one with a low accuracy. We used a Convolutional Neural Network architecture for MNIST and CIFAR10 and the VGG16 architecture for the Chest X-Ray images. The accuracies of the models are shown in Table~\ref{tab:accuracies}. 
\begin{table}
\caption{Accuracies of well-trained and badly-trained models for MNIST, CIFAR10, and Chest X-Ray images. The badly-trained models are trained with the same parameters but only on 1 or 2 epochs of the training set so that they have a low accuracy.}
\label{tab:accuracies}
\centering
\begin{tabular}{|c|c||c||c|}
\hline
& MNIST & CIFAR10 & Chest X-Ray \\
\hline
Well-trained & $99.1\%$ & $83.9\%$ & $94.5\%$ \\
Badly-trained & $69.02\%$ & $56.9\%$ & $76.5\%$ \\
\hline
\end{tabular}
\end{table}

We list our evaluation objectives below.
\begin{itemize}
    \item We demonstrate the need to define similar images that do not rely on adversarial perturbations or are only being visually similar to the original images.
    \item We use our metric to evaluate state-of-the-art attribution methods on the same networks and verify whether our observations about their performance align with the ones from the literature.
    \item We compare quantitatively our robustness metric to the current metrics of attributional robustness~\cite{bhatt2020evaluating,hsieh2020evaluations} as well as the fidelity metric, on networks with different accuracies trained on the same dataset to demonstrate the bias that previous robustness metrics introduce in their evaluation.
\end{itemize}

We evaluate our metric on the following well-known attribution methods: Saliency~\cite{simonyan2013deep}, DeepLift~\cite{shrikumar2017learning}, Integrated Gradients~\cite{sundararajan2017axiomatic}, Smooth Grad~\cite{smilkov2017smoothgrad}, Input $\times$ Gradient~\cite{shrikumar2016not}, Deconvolution~\cite{mahendran2016salient}, Layer-wise Relevance Propagation (LRP)~\cite{bach2015pixel}, and GreedyAS~\cite{hsieh2020evaluations}.

In practice, as suggested in~\cite{bhatt2020evaluating}, we also fix the size of the subsets used during the computation of the fidelity metric of an attribution method. More specifically, we use a subset of $(28*28)/2=392$ features for MNIST, $(32*32)/2=512$ features for CIFAR10, and $1500$ features for Chest X-Ray. The Chest X-ray images are RGB images of size $224\;\times\;224$, and hence, we don't use more than $1500$ features in the subset to be perturbed to ensure that the computation finishes within a reasonable time. However, fidelity is defined as the correlation between the importance of all possible different subsets of features and their effect on the model's output. Therefore, fidelity is time-consuming to compute exactly and use as an evaluation metric in practice. We randomly select the features of the input to be perturbed, that is, the features that will be set to a baseline value. As in~\cite{bhatt2020evaluating}, the baseline image for MNIST is a black image, while the baseline image for CIFAR10 and Chest X-ray images is the mean of the images in the test set. 

Finally, computing the Robustness $S_r$ metric or the attributions from the corresponding method GreedyAS is a time-consuming process that does not terminate within hours until we decrease the default number of samples that the method uses. The method could not be used for evaluating the larger RGB $224 \times 224$ Chest X-ray images. 
\subsection{Training of the GAN}
As discussed in Section~\ref{sec:motivating}, the GAN is trained on a dataset comprising approximately 1500 copies of the given image of interest, denoted as $\bm{x}$. Therefore, there is no requirement for an external dataset. By utilizing copies of $\bm{x}$ as the dataset, the requirement to establish bounds within the input space to ensure the generated images remain close to $\bm{x}$ is eliminated. Thus, the GAN generates images visually and numerically close to the original images, as they already serve as replicas of $\bm{x}$. Lastly, the training duration of the GAN is about 46 seconds for images of size $32 \times 32 \times 3$ (CIFAR10), and approximately 3 minutes for images of size $224 \times 224 \times 3$ such as the Chest X-ray images. The GAN architecture comprises 6 linear layers starting with 128 hidden neurons that are doubled in each layer with leaky ReLU activations and dropout layers in between. The input vector of the GAN is of size 100 and the output is of size ($32 \times 32 \times 3$) for CIFAR10 and ($224 \times 224 \times 3$) for Chest X-ray images.

\begin{figure*}[b]
    \centering
    \subfigure[][]{
        \label{fig:roc-a}
        \includegraphics[height=1.7in]{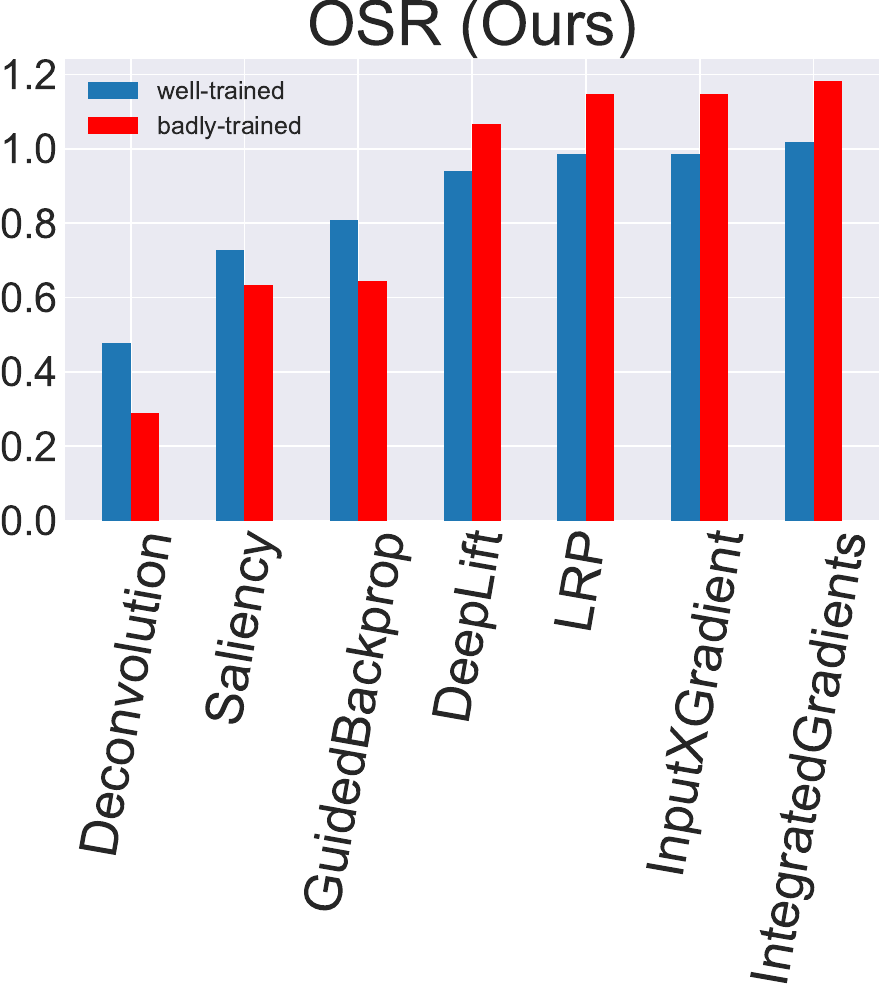}
    }
    \subfigure[][]{
        \label{fig:roc-c}
        \includegraphics[height=1.7in]{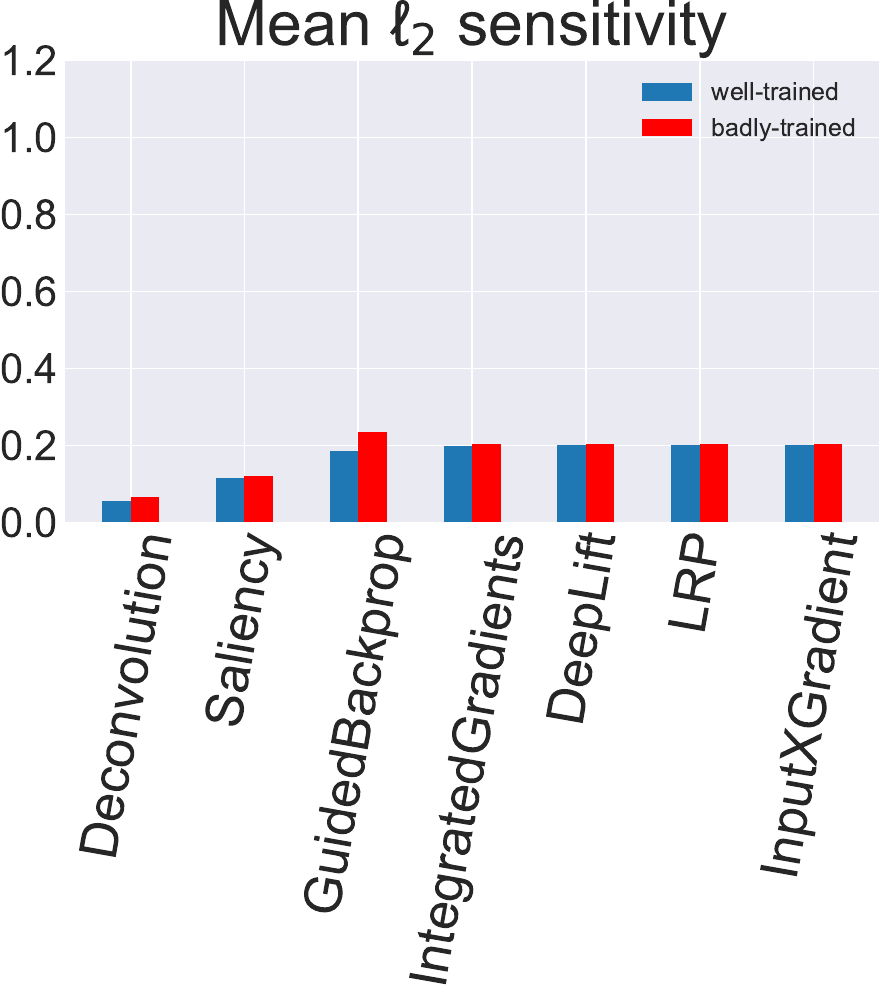}
    }
    \subfigure[][]{
        \label{fig:roc-e}
        \includegraphics[height=1.7in]{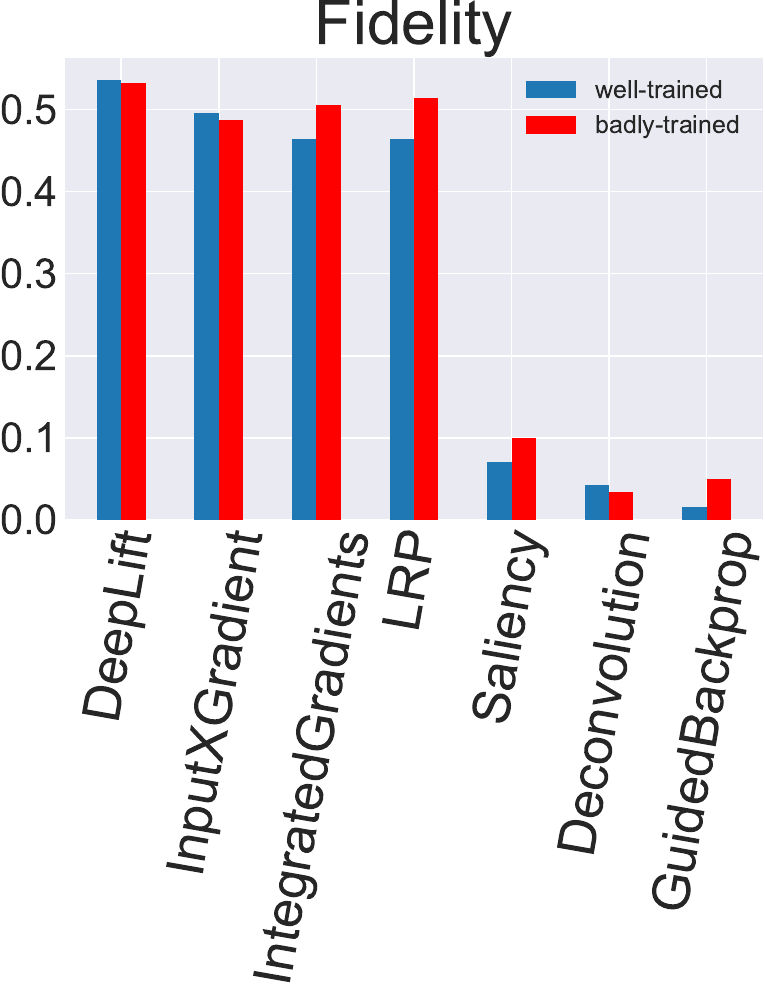}
    }
    \caption[Attribution Robustness and Fidelity.]{Comparison of attributional robustness metrics and fidelity on inputs of well-trained (\textbf{blue}) and badly-trained (\textbf{red}) networks for ChestXray images:
    \subref{fig:roc-a} OSR (Ours): Lower is better;
    \subref{fig:roc-c} Mean $\ell_\infty$ sensitivity $\mu_A$: Lower is better;
    \subref{fig:roc-e} Fidelity $\mu_F$: Higher is better.}
    \label{fig:robustness_metrics_covid}
\end{figure*}

\subsection{Evaluation of Similar Inputs}
First, we visualize the similar inputs produced by our GAN-based approach. 
Indeed, Figures~\ref{fig:mnist_gan}, \ref{fig:cifar10_gan}, and \ref{fig:covid_gan} show samples of generated images that have logits no more than $\delta$ apart from the original image's prediction logit. The GAN was trained to generate images similar to the image that is depicted in the first column of the Figures. It's crucial to note that each GAN generates images that exhibit visual similarity to the original image although not necessarily numerical similarity regarding their $\ell_2$ norms. Across all illustrations, it's evident that images produced by the GAN are visually alike and may not necessarily possess close $\ell_2$ distances. To provide context, the $\ell_{\infty}$ distances between the generated CIFAR10 images in Fig.\ref{fig:cifar10_gan} and the original image are shown in Fig.\ref{fig:eval-b}. This depiction also shows various $\ell_\infty$ distances between the generated images and the original. Furthermore, it's observed that as the discrepancy in prediction logit widens, the quality of the image diminishes. 

\subsection{Attributional Robustness with Neural Networks of Different Accuracy}
\begin{figure*}
    \centering
    \subfigure[][]{
        \label{fig:roc-a}
        \includegraphics[height=1.9in]{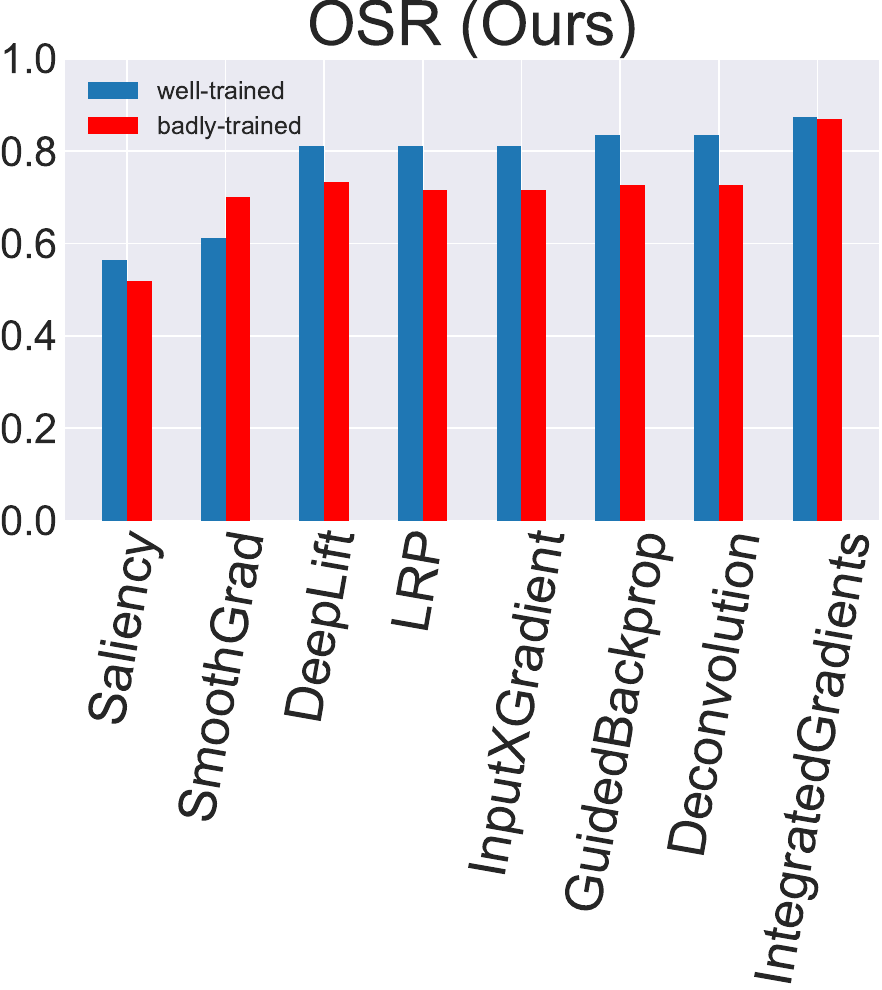}
    }
    \subfigure[][]{
        \label{fig:roc-c}
        \includegraphics[height=1.9in]{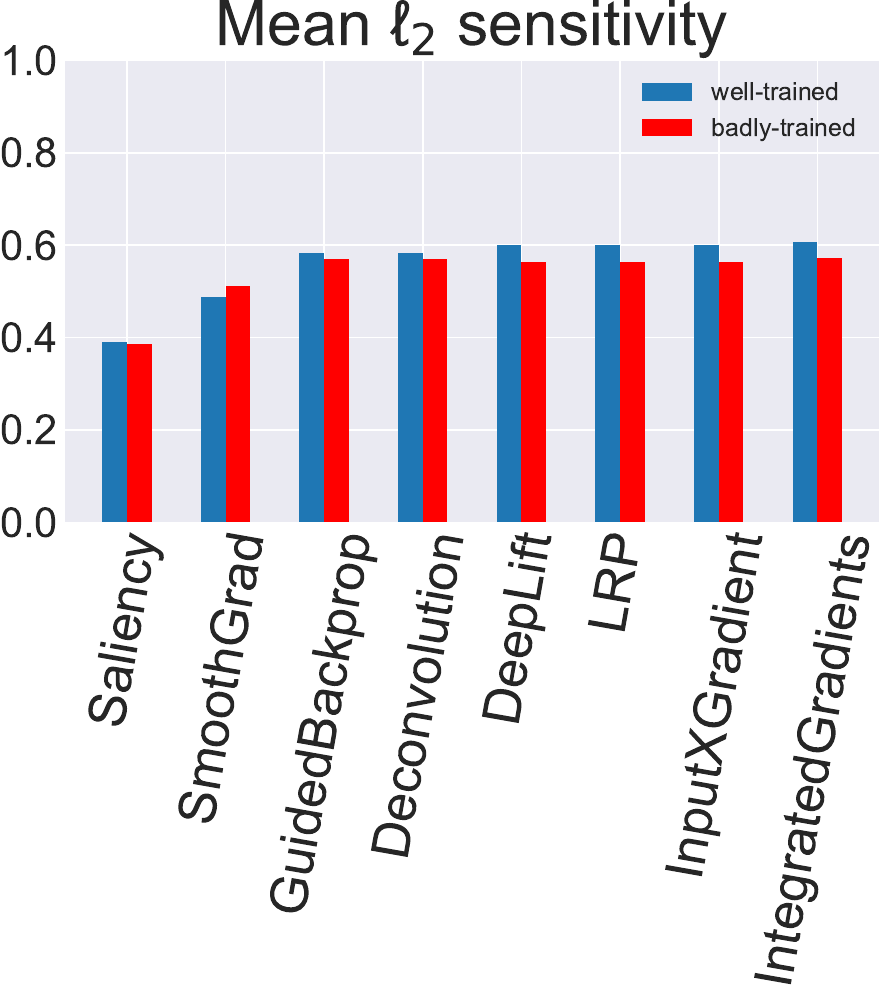}
    }
    \subfigure[][]{
        \label{fig:roc-d}
        \includegraphics[height=1.9in]{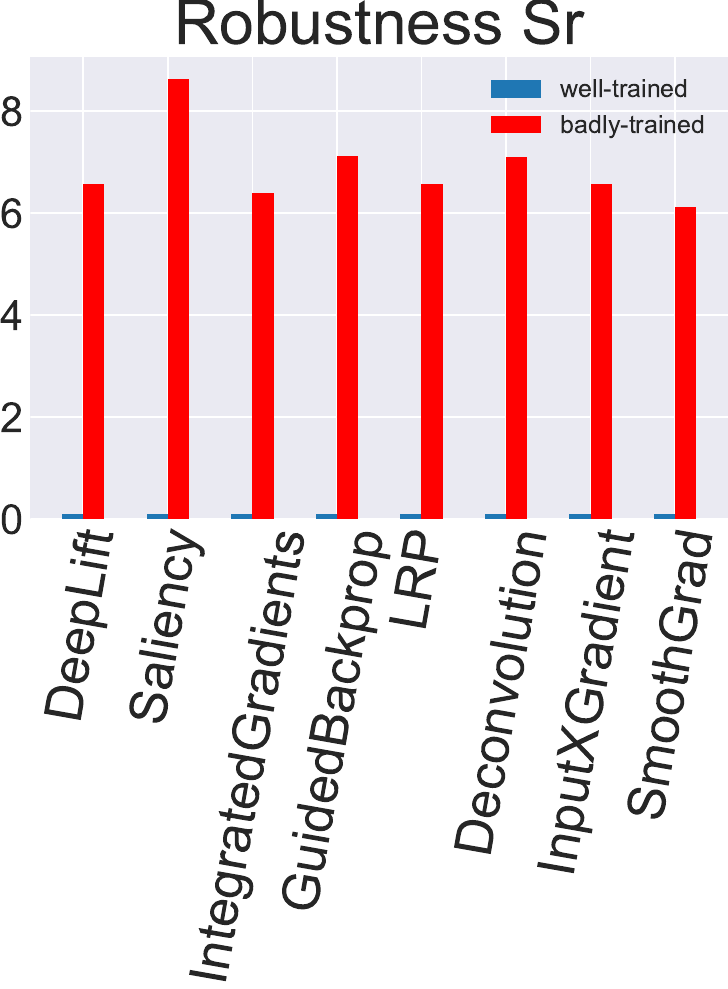}
    }
    \subfigure[][]{
        \label{fig:roc-e}
        \includegraphics[height=1.9in]{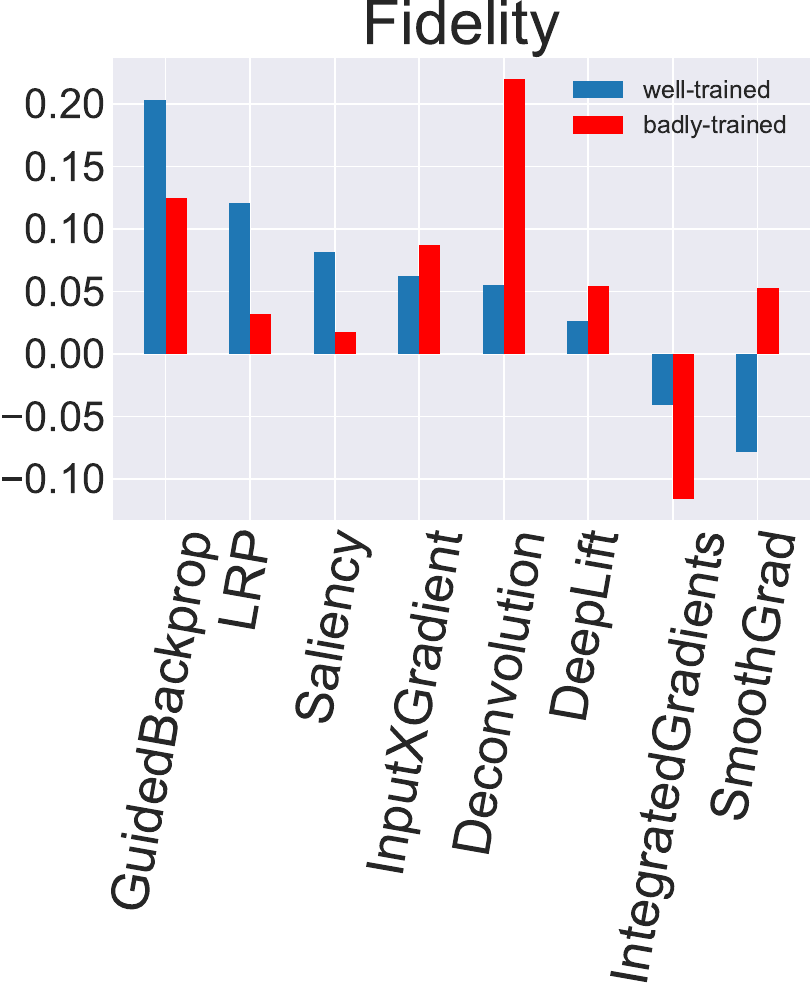}
    }
    \caption[Attribution Robustness and Fidelity.]{Comparison of attributional robustness metrics and fidelity on well-trained (\textbf{blue}) and badly-trained inputs (\textbf{red}) for CIFAR10:
    \subref{fig:roc-a} OSR (Ours): Lower is better;
    \subref{fig:roc-c} Mean $\ell_\infty$ sensitivity $\mu_A$: Lower is better;
    \subref{fig:roc-d} Robustness $S_r$: Lower is better;
    \subref{fig:roc-e} Fidelity $\mu_F$: Higher is better.}
    \label{fig:robustness_metrics_cifar10}
\end{figure*}

In Figures~\ref{fig:robustness_metrics_mnist}, \ref{fig:robustness_metrics_covid}, and \ref{fig:robustness_metrics_cifar10}, the results derived from a well-trained model regarding our metric OSR, sensitivity $\mu$, robustness $S_r$, and fidelity are represented using the blue color. Across all subfigures, the attribution methods are arranged based on the resulting order that the metric suggests for the well-trained model. For example, lower values of OSR or sensitivity indicate greater robustness in attributions. Likewise, lower values of Robustness-$S_r$ indicate better performance, while higher fidelity values signify better results.

Additionally, red is used to illustrate the results of the metrics on models trained to exhibit lower accuracies. This experiment aims to demonstrate how a neural network's accuracy reflects the distribution of inputs it has learned per class. Consequently, we expect the sensitivity metric to rank the robustness of attribution methods differently between well-trained and badly-trained models. This discrepancy arises because sensitivity evaluates similar inputs based on the dataset distribution rather than taking into account the neural network's learned input distribution.

Across all figures, we observe differences in the robustness shown by previously proposed metrics between well-trained and poorly-trained models. For instance, Smooth Grad, which ranks as the second-best robust method for the well-trained MNIST model, takes the top rank in robustness for the poorly-trained MNIST model compared to Saliency. Additionally, sensitivity places Integrated Gradients at the bottom for the well-trained CIFAR10 model, whereas it ranks as the fourth most robust method for the corresponding poorly-trained model. Similarly, sensitivity positions Guided Backpropagation as the third most robust method for the well-trained Chest X-ray model, while ranking it last for the poorly-trained counterpart.

Conversely, our metric consistently ranks the attribution methods in the same order for attributional robustness, irrespective of the model's training quality on the dataset.

Additionally, we observe a substantial difference in the robustness metric, $S_r$, between well-trained and poorly-trained models. $S_r$ is significantly lower for well-trained models compared to poorly-trained ones. This metric relies on identifying the minimal adversarial perturbation necessary to induce misclassification, thus attributing significance to the perturbed features accordingly. Consequently, $S_r$ is heavily influenced by the model's adversarial robustness.

Finally, we use fidelity to evaluate how faithful or accurate attributions are in comparison to their robustness evaluation. Fidelity deems methods with low robustness, such as Integrated Gradients and Input $\times$ Gradient, as being more faithful to the neural network's decisions, i.e., possessing higher fidelity. Interestingly, Smooth Grad, despite being considered the second most robust method, exhibits the lowest fidelity. This raises the question of whether there is a trade-off between robustness and fidelity in a trained neural network.

We conclude that our metric offers a more suitable approach towards establishing an objective evaluation of attribution methods. Prior methods have primarily relied on adversarial attacks to construct evaluation testbeds for attributions, leading to biases in the evaluation process. Subsequent research works have taken into account these attacks to introduce regularizers during training, aiming to enhance the robustness of attributions against the current testbed, which includes these adversarially attacked inputs.

%% file: sections/conclusion.tex
Our analysis underlines the need for a more objective metric capable of pinpointing the shortcomings of attribution methods rather than those of the neural network itself. Our robustness metric represents a way of distinguishing wrong explanation outcomes resulting from an attribution method from accurate explanations of poorly trained models. This shift allows for the evaluation of the explainability method's robustness independently of the neural network's robustness. Furthermore, our robustness metric offers guidance to end-users in selecting suitable attribution methods and assessing their robustness.

As feature attribution methods gain traction for explaining neural networks and new methods are introduced, there arises a need for proper evaluation of their robustness. This is crucial to avoid conflating weaknesses of the methods with weaknesses of the trained model and vice versa. In our paper, we introduce the first metric that aims to address this challenge. Our new robustness metric does not rely solely on adversarial inputs to compare and evaluate attribution methods. Our findings encourage the use and further development of our metric, further efforts to disentangle attributional robustness from adversarial robustness and explore the relationship between bounded inputs and bounded outputs concerning similarity for attributional robustness.